# Table of Content Detection using Machine Learning: Proposed System

Rachana Parikh[1] and Prof. Avani vasant[2]

[1] PG Student, C.E Dept., V.V.P. Engineering College, Rajkot, G.T.U
rachnaparikh88@gmail.com
[2] H.O.D, I.T Dept., V.V.P Engineering College, Rajkot, G.T.U
avanivasant@yahoo.com

## ABSTRACT

*Table of content (TOC) detection has drawn attention now a day because it plays an important role in digitization of multi-page document. Generally book document is multi-page document. So it becomes necessary to detect Table of Content page for easy navigation of multi-page document and also to make information retrieval faster for desirable data from the multipage document. All the Table of content pages follow the different layout, different way of presenting the contents of the document like chapter, section, subsection etc. This paper introduces a new method to detect Table of content using machine learning technique with different features. With the main aim to detect Table of Content pages is to structure the document according to their contents.*

## 1. INTRODUCTION

As today's era turns to digital from the physical object, all the documents like news papers, magazines and book documents are available digitally from digital library and it becomes a common need of all of us. All the books available digitally, are formatted with table of content for their content organization. In other words TOC is simply collection of references to individual articles of a document no matter what layout it follows.

Table of content includes the titles or description of section, and subsection. Along with that it also shows the order in which the content appear. It is a great way of organizing content of multipage document. It is also easy way for users to navigate the pages of document.

For easy retrieval of the intended portions as demanded by users, users need to go through table of content page. So it requires detecting the table of content page. Generally all the table of Content pages has different layout structure. So detection of Table of content pages is very exciting. TOC detection can be applied to the magazine, digital library and also to the book documents with different method and different algorithms available.

Based on general observed feature for detecting TOC pages are follows:

1. Page number-related heuristics and page Images.
2. Chapter, section and subsection are the structural element used to extract the Indentation and font size of multi-page document.

3. Predefined connector such as dot lines.
4. Headline and decorative element matching.
5. For Chinese book geometrical rules and semantic rules are combined.
6. Labeling approach to delimit articles.
7. Probabilistic relaxation through training on the layout of several representative samples.

In this paper we have proposed a system to detect table of content pages using machine learning approach. A decision tree algorithm is applied for table of content detection. The main aim is to achieve high accuracy for detection.

## 2. LITERATURE REVIEW

There are various techniques available for the Table of content detection.TOC can be detected with different approach with different features. Different approach can be combined for new approach discovery. For the discovery of new approach, different approaches can be combined.

S.Mandal et al[1] proposes a method of TOC detection based on morphology based algorithm. Here the first step of identification of TOC based on finding page numbers associated with the name of sections, sub-sections or chapter. In this method first step is to find word halos then finding the right-most word and then TOC is detected. This method is based on finding the page number and other text layouts using decision tree.

Adna Ul-Hasan et al[2] proposes a method of TOC detection is based on OCR free detection. The first step is to binarized document image. Then after digits and non-digit are extracted using MLP classifiers. Then distribution of digits on a page is estimated by vertical projection. A typical TOC page exhibits a vertical peak in projection profile. The width of projection profile is then determined to decide whether the input page is a TOC or not.

Shinji Tsuruoka et al[3] proposes a method that consists of pre-processing, structure understanding and final is to convert to xml formate.xml generation is necessary to understand structure of document pages.

## 3. APPLICATIONS OF TOC DETECTION

 Adna Ul-Hasan et al [2] shows that the identified TOC is useful for improving the accuracy of other document understanding techniques, for example:

## 1.  Advertisement Detection

The identified TOC can be fed to the advertisement finder component to identify advertisement pages .In this type of application TOC pages can be used to:

a) Mark pages that appear in the TOC as article page,
b) Generally pages, before TOC page are advertisement pages.

## 2.  Title Detection

A TOC page provides the way to find the article title from the text of TOC pages. Generally most significant title i.e. editor wants to use to attract the reader's attention is placed in TOC page.

**3. For easy retrieval of information as demanded by user.**

As user want to retrieve specific information we can locate that information from Table of Content page. So retrieval of demanded information becomes easier.

**4. Navigation to specific page document is also become easy and fast.**

As user want to retrieve specific information from the specific page number, navigation to that specific page number becomes easy. All the topics are organized in Table of content page with page number.

**5. Table of content pages can also be detected from magazines and journals.**

Table of content pages can also be detected from the different multipage document which follows different layout of content. e.g. magazine and journal have different layout than book document.

**6. Avoiding Ctrl-F for finding specific topic from multi-page document.**

Generally in e-books Table of content are searched by ctrl-F.so whenever it is required to search content, we need to go though ctrl-F for searching, which is disabled in some documents that are mere scanned copies. So proposed system provides an easy way to search content automatically.

**7. For custom publication of the book we have to detect ToC page for customized the book.**

Custom publication provides the facility to merge the different module from different book. So for merging the chapter we need to detect table of content page.

## 4. PROBLEM STATEMENT

Main aim is to detect Table of Contents automatically. For automatic detection of Table of Content, we apply machine learning Technique. Specifically Decision tree algorithm will be applied on e-books or digital libraries. Finally we calculate the overall accuracy using various attributes like presence of title term, font type of title term, Font class of Title term, Number of contextual term, Normalized frequency of Toc section term, Normalized frequency of line start with number, Normalized frequency of line end with number, Are all the English numbers in ascending order and normalized line position of title term.

Many different approaches are available to identify the TOC pages of multi-page document. Majority of the approach are for specific formats features to identify TOC pages for specific class of document only. Like decorative element such as page numbering, chapter, section and subsection numbering formats or indentation detection. Some of the approaches are language specific. So varieties of different layouts are available. As TOC pages are free from specific format. It is challenging task to detect TOC pages having different layout formats [6].

## 5. CHARACTERISTICS OF TOC PAGE

1. Generally TOCs' are available as a full page in books and report. It also may be a part of a page; some    journals TOCs are available with logo and headings and all other texts.

2. Generally rightmost columns contain the digits i.e. page numbers. But it is not necessary that this format is always maintained in each document. It may also be possible that page number appears just after the name of the sections and subsection headings.
3. Leftmost columns usually have the section and subsection numbers. The world gap between the title and section number is much more than the normal word gaps.
4. Dotted lines or thin lines appear at regular intervals in the TOC pages. This dotted lines or thin line show the correlations between the end of the text line and the corresponding page numbers at the right-most columns.
5. TOC page may have multiline title. Many TOCs are available with horizontal separator lines.
6. TOCs have characteristics of similar to the tabular Structure.
7. Page numbers are short mostly 3 for average books[5]

## 6. PROPOSED APPROACH

As shown in fig-1 there are various approach available for machine learning technique for implementing the system. All the chosen approaches are described below.

### A. Machine Learning

Machine learning is define as an art of developing computer programs in such a way that can teach themselves to grow and change when exposed to new data without being explicitly programmed.

The work is to detection of ToC pages automatically so machine learning provides the efficient way for automation.

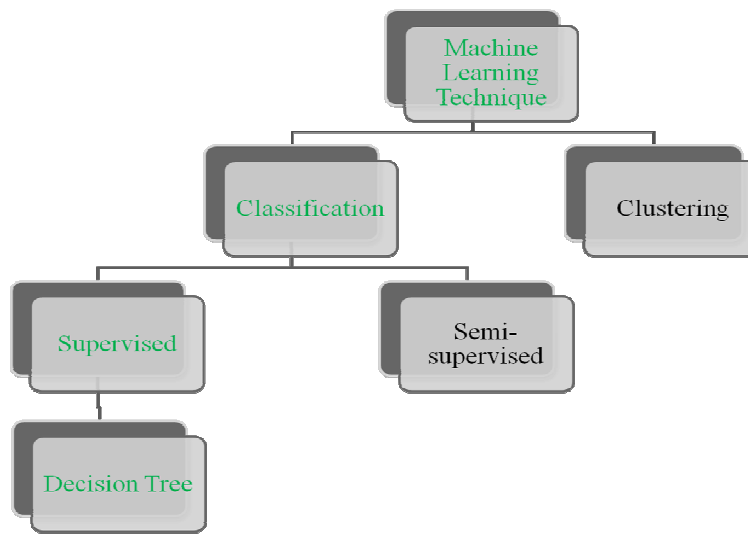

Fig-1 Flow of Proposed approach

### A Classification

Classification is a task of assigning class label to set of unclassified cases, on the basis of training set.

Classification is a formal method for repeatedly making judgments in new situations.

Here work is to detect ToC pages from e-book, so pages will be classified into ToC and Non-ToC pages.

**B Supervised Learning**

Supervised learning means the results or the possible outcomes are already known in advanced from the training set of data.

Here the work will be done with first training the system and then system will learn by itself. So, in training process the possible outcomes will be well known in advance.

**C  Decision Tree**

Decision tree is a structure that can be used to divide up a large collection of records into successively smaller sets of records by applying a sequence of simple decision rules.
Decision tree provides human interpretability so it will be easy to train the system.

## 7. FLOW OF PROPOSED SYSTEM

1.   Having digital library with offline pdf's or another document.
2.   Choose 15-30% pages of single document pages.
3.   Labeled TOC and NON-TOC accordingly.
4.   Train the system with various attribute of the page layout.
5.   Feed some new document to test the system
6.   Observe the learned data for system performance.

As shown in fig-2 proposed system uses the digital library or offline pdf's or any multipage document as input. For the Table of content detection first 15-30% pages are selected from the entire document and given to the system as input. All the selected pages are labeled as ToC and NON-ToC accordingly. Labelling is done manually for the training of the system correctly. Train the system with different attribute of table of content detection. Table of content pages have several different attributes.

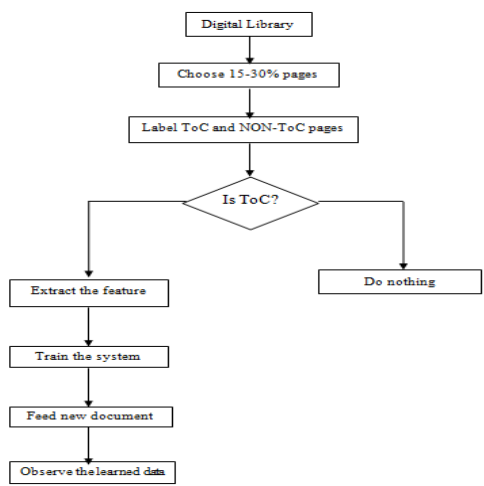

Fig-2 Flow of Proposed system

All the Table of Content pages follow different layout for presenting their content. With different attribute train the system for the detection. In other words training is done by labelling .So system can easily learn from the training i.e. from labelled data. After learning from labelled data system will be tested for correct detection. Observe the input and output data for system testing. Match the numbers of correctly detected and actual Table of content pages for the performance of the system.

## 8. PROPOSED ALGORITHM

Step-1:-Take some pdf or multipage document for detecting ToC pages.
Step-2 :-Convert Pdf or multipage document to XML.
Step-3:- Read the XML file to extract the feature.
Step-4:- Extract the following feature to detect the ToC page
      I.    Presence of ToC Title term.
     II.    Font type of ToC Title term.
    III.    Font class of ToC Title term.
    IV.    Number of Contextual terms.
     V.    Number of section term.
    VI.    Line position of Title term.
   VII.    Line start with number.
  VIII.    Line end with number.
    IX.    Are English numbers in ascending order?

Step-5:-Construct an output file with the extracted features from document for decision tree generation.
Step-6:-Label the training data with class label (ToC/NONToC).
Step-7:-Learn a decision tree out of the training data.
Step-8:-For unseen data
      (a)Follow step-1 to step-6
      (b)Apply decision tree learning step-7 to find class labels.

Here some pdf or multipage documents that are with table of content page is selected for Table of content page detection. For detection of table of content page we need to extract the features from the pdf or multipage document. For extracting feature we need to convert document in structured format. So selected document is converted into XML formate.XML provides the structured format with different layout related attribute. Converted Xml is used to read the data contained in that document.

There are many features available for the detection of table of content page from the multipage document. Layout based, number based and term based attributes. Proposed system uses mixture of all the type of attribute that are term based and number based. The selected attribute for detection of table of content are as follows.

1. Presence of title term means need to check for the title term i.e. Table of Content or Content word. Most of the tables of content pages are with term Table of content or content word only.
2. If title term exists then type of title term will be different then other words.  Most of the title terms are in Bold. So it can be easily differentiable.
3. Same as font type, title term has different class of font. Generally it is in Times New Roman. It may also differ document to document.

4. For detecting the actual title term, need to look for the neighbors of the title term i.e. contextual terms. If its successor and predecessor does not exists then it is title term only.

5. Normalized frequency of section term. It means frequency of section term is higher than the possibility of page is higher for the table of content page.

6. Line position of the title term means if title term exists then need to look for the line number in which line the content word exits. If title term exits in starting line then that page is title term otherwise not.

7. Normalized frequency of line start with number means subsection have points indicated by number so it can be easily be shown that it may be title term. E.g.1.1, 1.2, 1.3 etc all are the number from which line can be started.

8. Normalized frequency of line end with number that means all the topics and subtopic display the page numbers. e.g.1,2, 3,4,5,6 etc.

9. Are English numbers in ascending order? Means all the English number i.e. all the page numbers and section numbers are always in ascending order.

After extracting the feature need to generate the output file with all the attribute values for decision tree generation. An output contains attribute values from which decision tree will be generated

Label the training data with class label as TOC or NON-TOC accordingly for the training the system. System will learn from the class labels. For unseen data, follow the entire procedure.

The proposed algorithm can be used for most of the different page layout. It provides the high accuracy of correctly detected Table of content pages. All the proposed features are unbiased for particular term or number. For the implementation of algorithm various tools are available.

## 9. Training dataset

Various attributes are used for detecting table of content page. following are the attributes used to train the system for table of content detection. These training set is sample training set for proposed system. According to the training set decision tree will be generated. from the decision tree we can easily understand that what system is learning.

**Table I: Training dataset**

| Contains TOCTitleTerms | StyleOfTOCTitleTerms | NormlisedFrequencyOfLineEndingWithNumber | NormalisedFrequencyOfOutgoingLinks | NormalisedFrequencyOfLineStartingWithNumber | Label |
|---|---|---|---|---|---|
| YES | LARGEST | 0.8 | 0.89 | 0.8 | TOC |
| YES | LARGEST | 0.1 | 0.56 | 0.05 | TOC |
| YES | INTERMEDIATE | 0.9 | 0.67 | 0.13 | TOC |
| YES | INTERMEDIATE | 0.2 | 0.96 | 0.18 | TOC |
| YES | MOST-FREQUENT | 0.86 | 0.91 | 0.83 | TOC |
| YES | MOST-FREQUENT | 0.13 | 0.85 | 0.07 | TOC |
| NO | NA | 0.98 | 0.91 | 0.12 | TOC |
| NO | NA | 0.16 | 0.87 | 0.103 | TOC |
| NO | NA | 0.87 | 0.3 | 0.02 | NON-TOC |
| YES | MOST-FREQUENT | 0.2 | 0.86 | 0.88 | NON-TOC |

From the training dataset values of the different features are different. Three of them uses the normalized frequency of the presence of the attribute or feature for detecting the table of content page. Rest of the attribute having the different values according. From the training set attribute value decision tree is generated.

## 10. Decision tree generation

Decision tree is generated automatically from the dataset. According to the dataset value decision tree may vary. System will learn from the training dataset and generate decision tree. After training the system, system will learn by itself from the training. So table of content is detected automatically by system. The decision tree for the given above dataset is shown below.

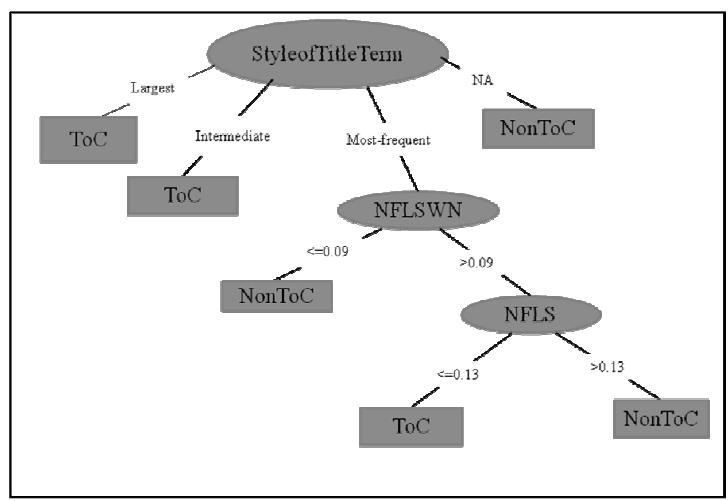

Fig-3 Decision tree for Dataset

As shown in decision tree all the attribute are set according their normalized frequency. Each and every time the decision tree is generated according to attribute value. Decision tree is human interpretable so that we can easily understand what machine is learning from our different inputs.

## 11. CONCLUSION

In this paper we have pros with posed an approach for TOC detection using decision tree algorithm with machine learning technique. For detecting Table of Content various attributes are to be derived like presence of title term, font type of title term, Font class of Title term, Number of contextual term, Normalized frequency of Toc section term, Normalized frequency of line start with number, Normalized frequency of line end with number, Are all the English numbers in ascending order and normalized line position of title term, all the attributes are also specified and explain. These attributes will be applied for table of content detection. The future work is to implement the proposed system and measure the accuracy of Table of content detection.

## Biography

**Rachana P. Parikh** is a PG scholar from Vyavasayi Vidya Pratishthan Engineering College, Gujarat Technological University, Gujarat, India. Her area of interest is Artificial Intelligence.rachnaparikh88@gmail.com 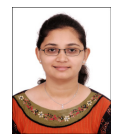

**Prof.Avani Vasant** is an assistant professor And Head of Information Technology department in Vyavsayi Vidya Pratishthn Engineering College, Rajkot, Gujarat, India. She is having more than 13 years of teaching Experience. Her area of Interest is Artificial Intelligence.avanivasant@yahoo.com 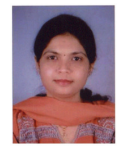